\documentclass{article}

\usepackage{spconf,graphicx,hyperref}
\usepackage{booktabs}
\usepackage{amsmath,amssymb}
\usepackage[inline]{enumitem}
\usepackage[accsupp]{axessibility}
\usepackage{xcolor}
\usepackage{multirow}
\usepackage{microtype}          % improves line-breaking & letter-spacing
\usepackage{placeins}           % provides \FloatBarrier
\usepackage{colortbl}           % colored table rules
\usepackage{tcolorbox}
\usepackage{xurl}
\usepackage{graphicx}
\usepackage[table]{xcolor}
\tcbuselibrary{listings,breakable}

\title{EgoPhys: Estimating Peak Contact Force and Mechanical Work from Egocentric Manipulation Video}
\name{
Zhuo Dong$^{1}$,
Jianhua Yang$^{2}$,
Haohao Li$^{3}$,
Yumeng Zhao$^{1}$,
Keji He$^{1}$,
Yan Huang$^{2}$,
Liang Wang$^{2}$
}

\address{
$^{1}$ School of Artificial Intelligence, Shandong University, Jinan, China \\
$^{2}$ Institute of Automation, Chinese Academy of Sciences, Beijing, China \\
$^{3}$ School of Mechanical Engineering, Tianjin University, Tianjin, China
}

\begin{document}
%\ninept
%
\maketitle
\begin{abstract}

%Manipulating articulated objects requires a robot to reason not only about how an object moves, but also about the physical loads needed to move it. The Hoi! dataset pairs head-mounted video with synchronized force and motion measurements, yet these physical signals are unavailable to a vision-only system at inference. We study the prediction of an instrumented gripper's peak resultant contact force and contact-phase mechanical work from egocentric RGB video. The task is challenging because object resistance is not directly visible and camera motion can obscure interaction motion. Moreover, peak force is often determined by a brief contact transient, whereas mechanical work integrates force along the motion over time. We propose an interaction-aware framework with semantic, event, and motion experts, which are routed independently for the two targets . On a fixed 36-clip outer-test split of Hoi!, our method achieves $5.205\pm0.584$ N force MAE and $0.894\pm0.081$ J work MAE.

%Understanding articulated-object manipulation requires reasoning about both object motion and the physical effort involved, which can be characterized by peak contact force and mechanical work. Estimating these quantities from egocentric video is challenging because physical interaction cues are local and indirect. 

Physically grounded manipulation of articulated objects requires understanding both the maximum forces encountered during contact and the work performed as their parts move. Peak contact force and mechanical work quantify these complementary aspects, but estimating them from egocentric video is challenging because physical interaction cues are local and indirect. Moreover, peak force is associated with brief contact events, whereas mechanical work depends on force-motion coupling throughout the contact duration. To address these challenges, we propose EgoPhys, an RGB-only framework comprising Contact-Aware Spatial Aggregation (CASA) and Target-Specific Multi-Expert Temporal Routing (TMTR). CASA integrates appearance and geometry features to emphasize interaction-relevant cues, while TMTR models semantic, event, and motion cues with specialized temporal experts and routes them separately for force and work prediction. On the test split from Hoi! dataset, EgoPhys substantially improves predictions of peak force and mechanical work, achieving MAEs of \(5.205 \pm 0.584\) $N$ and $0.894 \pm 0.081$ $J$, respectively.

\end{abstract}

\section{Introduction}
\label{sec:introduction}

Understanding and manipulating articulated objects, such as drawers and doors, requires reasoning about how their parts can move and the forces needed to move them. Existing vision-based methods localize actionable regions \cite{mo2021where2act}, infer articulation geometry and motion \cite{osti_10400254,zhang2023flowbot++}, or learn manipulation policies \cite{wang2025articubot}. These methods characterize articulated part motion, but provide limited information about the physical quantities involved in interactions.

%These methods primarily characterize whether and how articulated parts can move, but provide limited information about the physical quantities associated with executing interactions. 

%These methods largely emphasize geometric and kinematic reasoning, while physical interaction quantities such as contact force and mechanical work have received comparatively less attention. 

\begin{figure}[t]
    \centering
    \includegraphics[width=\columnwidth]{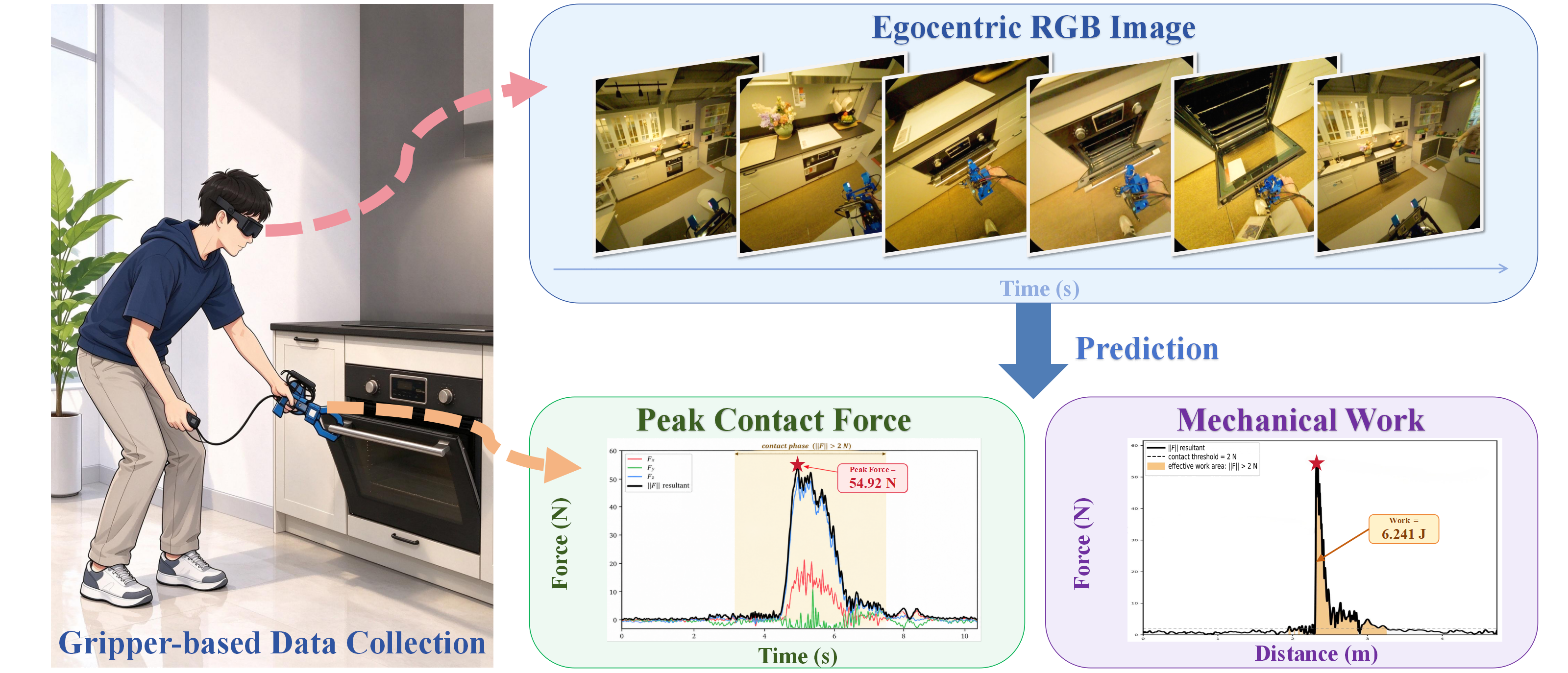}
    \vspace{-0.6cm}
    \caption{Task overview. Egocentric RGB video is captured by a head-mounted camera, while force/torque and tool-motion signals are recorded by the Hoi! custom gripper. The task is to predict peak contact force and contact-phase mechanical work from the egocentric RGB video alone. }
    \label{fig:hoi-capture}
    \vspace{-0.4cm}
\end{figure}

%Previous studies have considered force estimation under several task formulations. 
Previous studies have explored different force estimation tasks. Pham et al. \cite{pham2015towards} and Li et al. \cite{li2019estimating} recover interaction forces from observed hand-object motion under contact and dynamics constraints. PressureVision++ \cite{grady2024pressurevision++} and EgoPressure \cite{zhao2025egopressure} estimate pressure at hand-object contacts from RGB observations, while ForceSight \cite{Collins2024ForceSight} predicts a three-dimensional force goal from an RGB-D observation and a language instruction. Vision-based tactile methods such as Sparsh \cite{higuera2024sparsh} and FeelAnyForce \cite{shahidzadeh2025feelanyforce} infer contact forces from local sensor deformation, where contact is directly observed. However, these formulations rely on explicit dynamics, measured depth, task instructions, or tactile observations. Estimating interaction force and work directly from egocentric RGB video remains comparatively underexplored.  

Egocentric video with synchronized force measurements remains scarce. HOI4D \cite{Liu2022HOI4D} and Ego4D \cite{Grauman2022Ego4D} capture rich egocentric interactions but lack force sensing, while RH20T \cite{fang2024rh20t} and ForceMimic \cite{liu2025forcemimic} record force or wrench signals mainly for robot learning. The custom-gripper recordings in Hoi! \cite{engelbracht2026hoi} bridge this gap by providing head-mounted egocentric video synchronized with force/torque and tool-motion measurements during articulated manipulation. Building on these synchronized signals, we study sequence-level prediction of peak contact force and contact-phase mechanical work from egocentric RGB video.

%Hoi! dataset \cite{engelbracht2026hoi} bridges this gap by providing synchronized egocentric video, synchronizing egocentric video with force/torque, tactile sensing, and tool motion during articulated manipulation. It also evaluates ForceSight \cite{Collins2024ForceSight}, which predicts a 3D force goal from a single RGB-D observation and a manipulation goal. This alignment provides the physical supervision needed to estimate interaction-level quantities from egocentric RGB video alone.

As illustrated in Fig. \ref{fig:hoi-capture}, we predict peak resultant contact force and mechanical work over the contact phase from a short head-mounted RGB clip. The task has two key challenges. First, the visual consequences of physical interaction are often subtle and spatially localized around the gripper–object contact region, and can therefore be overwhelmed by background content and egocentric camera motion. Second, the two targets exhibit different temporal dependencies: peak force is governed by brief high-load events, whereas mechanical work depends on force–motion coupling throughout the contact phase. These characteristics call for interaction-centered spatial aggregation and target-specific temporal modeling.

%First, the most direct force-related cues are local. They occur around the gripper-object interaction and can be diluted by the rest of the frame. Second, peak force and work accumulate force along the interaction motion. Scene semantics provide additional context for both. These characteristics call for interaction-centered spatial aggregation and target-specific temporal modeling.

%These observations call for interaction-centered spatial modeling and target-specific temporal modeling.

\begin{figure*}[th]
    \centering
    \includegraphics[width=0.82\linewidth]{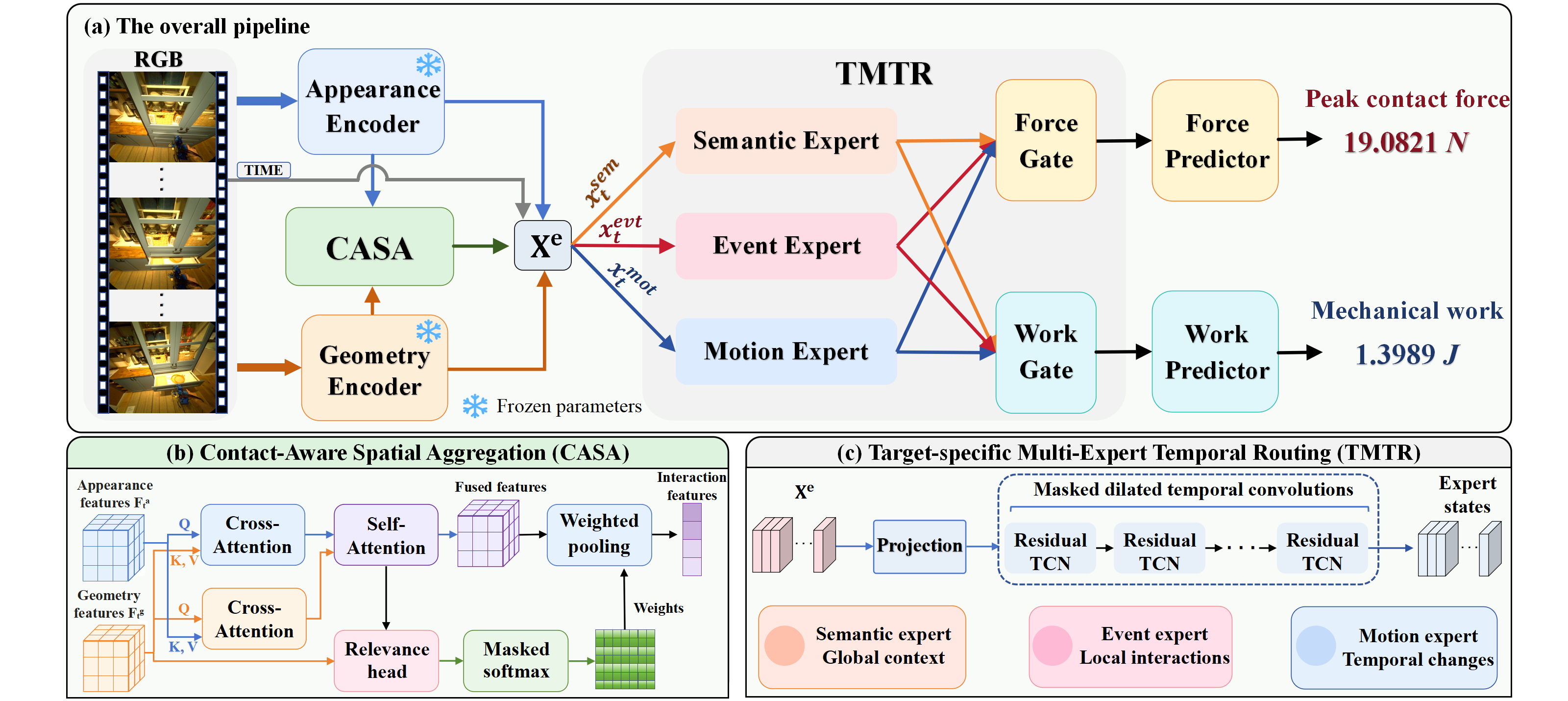}
    \vspace{-0.3cm}
    \caption{Overview of our proposed EgoPhys framework. Given an egocentric RGB clip, frozen appearance and geometry encoders extract complementary visual features. CASA aggregates interaction-relevant spatial evidence, while TMTR models complementary temporal cues to predict peak contact force and contact-phase mechanical work.}
    \label{fig:method-overview}
    \vspace{-0.4cm}
\end{figure*}

To address these challenges, we propose EgoPhys with two main components. Contact-Aware Spatial Aggregation (CASA) integrates dense appearance and geometry features and selectively aggregates interaction-relevant spatial evidence. TMTR models semantic context, transient interaction events, and multi-scale motion changes with specialized temporal experts, and learns target-specific mixtures of their representations for force and work prediction. Based on the algorithm described above, we won the ECCV 2026 Hoi! Force Prediction Challenge \cite{hoi_challenge2026}. 
%learns spatial weights to preserve interaction-relevant evidence. Target-Specific Multi-Expert Temporal Routing (TMTR) models semantic, event, and motion streams with separate temporal experts and combines them through independent gates for force and work. 

Our contributions are threefold: (\textbf{i}) We investigate joint estimation of peak contact force and contact-phase mechanical work from egocentric RGB video and propose EgoPhys, an RGB-only framework. (\textbf{ii}) We introduce CASA and TMTR to aggregate interaction-relevant evidence and learn target-specific mixtures of semantic, event, and motion representations. (\textbf{iii}) Experiments on a fixed Hoi! test split demonstrate the effectiveness of EgoPhys for both targets.

%Our contributions are threefold: (\textbf{i}) We investigate the joint estimation of peak contact force and contact-phase mechanical work from egocentric RGB video and propose EgoPhys, an RGB-only framework for the task. (\textbf{ii}) We introduce CASA and TMTR to aggregate interaction-relevant evidence and learn target-specific mixtures of semantic, event, and motion representations; (\textbf{iii}) We evaluate EgoPhys on a fixed test split of the Hoi! dataset, demonstrating its effectiveness in force and work estimation.

%We evaluate EgoPhys on a fixed test split constructed from Hoi! dataset, demonstrating the effectiveness of the proposed method on force and work estimation. 

%CASA, which integrates appearance and geometry while preserving local interaction evidence; (\textbf{ii}) TMTR, which allows peak force and mechanical work to use different mixtures of semantic, event, and motion cues; and (\textbf{iii}) an evaluation on Hoi!, where EgoPhys achieves \(5.205 \pm 0.584\) N MAE for peak contact force and \(0.894 \pm 0.081\) J MAE for contact-phase mechanical work on the fixed 36-clip outer-test split.

\section{Method}
\label{sec:method}

\subsection{Problem Formulation}

Given a short head-mounted RGB clip $\mathcal{X} = \{I_{t}\}_{t=1}^{T}$, our goal is to predict peak contact force and mechanical work for the gripper-object interaction. Following the force estimation setting in the challenge of the Force-Grounded, Cross-View Articulated Manipulation \cite{hoi_challenge2026}, the targets are derived from synchronized gripper force/torque measurements and end-effector velocity. Let $\Delta$ denote the interaction interval and $\mathcal{C}=\left\{t\in\Delta: \|\mathbb{F}(t)\|_2>2\,\mathrm{N} \right\}$ denote the contact set. The two targets are defined as
\begin{equation}
P_f=\max_{t\in\Delta}\|\mathbb{F}(t)\|_2 ;
\qquad
P_w=
\left|
\int_{\mathcal{C}}
\mathbb{F}(t)^{\top}\mathbf{v}(t)\,\mathrm{d}t
\right|,
\end{equation}
where $\mathbb{F}(t)$ and $\mathbf{v}(t)$ denote the contact-force vector and the velocity of the end-effector at time $t$, respectively. $P_{f}$ measures peak resultant contact force (\textit{Newtons}), while $P_{w}$ measures the magnitude of work during contact (\textit{Joules}).

\subsection{Framework Overview}

As illustrated in Fig. \ref{fig:method-overview}, EgoPhys consists of Contact-Aware Spatial Aggregation (CASA) and Target-Specific Multi-Expert Temporal Routing (TMTR). CASA fuses appearance and geometry features and selectively aggregates interaction-relevant spatial evidence. TMTR models semantic, event, and motion streams with separate temporal experts and routes them independently for force and work prediction.

\subsection{Contact-Aware Spatial Aggregation}

CASA integrates complementary appearance and geometry cues while emphasizing spatial regions relevant to physical interaction. For frame $t$, the dense features extracted by DINOv3 \cite{simeoni2025dinov3} and Depth Anything 3 (DA3) \cite{lin2025depth} are projected into a common space and denoted as $\mathbf{F}_t^{\mathrm{a}},\mathbf{F}_t^{\mathrm{g}} \in\mathbb{R}^{B\times P\times D}$, where $P=hw$ is the number of spatial locations and $D$ is the embedding dimension. Then, we utilize bidirectional cross-attention to exchange information between them:
\begin{equation}
\begin{aligned}
\widetilde{\mathbf{F}}_t^{\mathrm{a}}
&=\operatorname{CA}\left(
\mathbf{F}_t^{\mathrm{a}},
\mathbf{F}_t^{\mathrm{g}},
\mathbf{F}_t^{\mathrm{g}}
\right),\\
\widetilde{\mathbf{F}}_t^{\mathrm{g}}
&=\operatorname{CA}\left(
\mathbf{F}_t^{\mathrm{g}},
\mathbf{F}_t^{\mathrm{a}},
\mathbf{F}_t^{\mathrm{a}}
\right),
\end{aligned}
\end{equation}
where $\operatorname{CA}(Q,K,V)$ denotes cross-attention. The updated features are concatenated and refined through one layer spatial self-attention:
\begin{equation}
\mathbf{F}_t
=\operatorname{SA}(
[\widetilde{\mathbf{F}}_t^{\mathrm{a}};
 \widetilde{\mathbf{F}}_t^{\mathrm{g}}]).
\end{equation}

To emphasize interaction-relevant regions, a single-layer MLP is used to predict spatial scores, \emph{i.e.}, $\mathit{s}_{t}=MLP(\mathbf{F}_{t})$, where $\mathit{s}_{t} \in \mathbb{R}^{B \times P}$. The scores are normalized across spatial locations and used to compute a weighted frame-level representation:
\begin{equation}
\begin{aligned}
\mathbf{z}_t =
\sum_{p=1}^{P}
\alpha_{t,p}\mathbf{F}_{t,p}; \qquad
\alpha_{t,p} =
\frac{\exp(s_{t,p})}
{\sum_{q=1}^{P}\exp(s_{t,q})}.
\end{aligned}
\end{equation}
The resulting feature $\mathbf{Z} = \{\mathbf{z}_t\}^{T}_{t=1} \in \mathbb{R}^{B \times T \times D}$ is then passed to the subsequent temporal modeling.

\subsection{Target-Specific Multi-Expert Temporal Routing}

%TMTR models complementary temporal evidence with semantic, event, and motion experts, and routes their output separately for force and work prediction. Experts share the same architecture, but use independent parameters and distinct input features. 

%TMTR captures complementary temporal evidence using semantic, event, and motion experts, whose outputs are adaptively fused through separate force and work gates. Experts share the same architecture, but have independent parameters and distinct input features. 

TMTR uses three independently parameterized experts with a shared architecture: the semantic expert provides interaction context for both targets, the event expert captures brief changes indicative of peak force, and the motion expert models sustained changes associated with mechanical work. Target-specific gates then combine these complementary cues separately for the prediction of force and work.

\noindent\textbf{Expert input construction.}
The semantic expert uses the projected DINOv3 class token $\mathbf{c}_{t}^{cls}$ and the average-pooled patch feature $\mathbf{g}_{t}$ to encode scene and object context. The event expert uses the output feature $\mathbf{z}_{t}$ of CASA, which preserves interaction-relevant evidence. The motion expert captures changes in both interaction-centered and global cues through multi-scale temporal differences: 
\begin{equation}
\boldsymbol{\delta}_t^{(\ell)}
=[\mathbf{z}_t-\mathbf{z}_{t-\ell};
  \mathbf{g}_t-\mathbf{g}_{t-\ell}],
\quad \ell\in\{1,4,8\}.
\end{equation}
Each difference compares the current frame with the frame $\ell$ steps earlier, with the unavailable differences set to zero. A temporal embedding $\mathbf{r}_t$ is appended to each input stream:
\begin{equation}
\begin{aligned}
\mathbf{x}_t^{\mathrm{sem}}
  &=[\mathbf{c}_t^{\mathrm{cls}};\mathbf{g}_t;\mathbf{r}_t];
&
\mathbf{x}_t^{\mathrm{evt}}
  &=[\mathbf{z}_t;\mathbf{r}_t];\\[-1pt]
\mathbf{x}_t^{\mathrm{mot}}
  &=[\boldsymbol{\delta}_t^{(1)};
     \boldsymbol{\delta}_t^{(4)};
     \boldsymbol{\delta}_t^{(8)};
     \mathbf{r}_t].
\end{aligned}
\label{eq:expert_inputs}
\end{equation}

\noindent\textbf{Modeling of temporal experts.}
Each expert projects its input to the same dimension $D$, followed by six residual temporal convolution blocks:  
\begin{equation}
\mathbf{H}^{e}=\mathcal{E}_{e}(\mathbf{X}^{e})
\in\mathbb{R}^{B \times T \times D},
\quad e\in \{\mathrm{sem},\mathrm{evt},\mathrm{mot}\}.
\end{equation}
Here, $\mathbf{X}^{e}=\{\mathbf{x}_{t}^{e}\}_{t=1}^{T}$, $sem$, $evt$, and $mot$ denote semantic expert, event expert, and motion expert, respectively.  Each block uses a depthwise temporal convolution of kernel size $3$, followed by pointwise channel expansion
and projection, with layer normalization, GELU, dropout, and a residual connection. Dilation rates $1, 2, 4, 8, 16, 32$ capture the temporal context on multiple scales.

Then, two independent gates combine outputs of three experts $\{\mathbf{h}^{sem}_{t}, \mathbf{h}^{evt}_{t}, \mathbf{h}^{mot}_{t}\}$ for force ($f$) and work ($w$). Given the concatenated feature $\mathbf{q}_{t}=[\mathbf{h}_{t}^{sem}, \mathbf{h}_{t}^{evt}, \mathbf{h}_{t}^{mot}]$, the task-specific representations are
\begin{equation}
\begin{aligned}
\boldsymbol{\pi}_t^k
&=softmax(
\mathbf{W}_{k} LN(\mathbf{q}_t)+\mathbf{b}_k); 
\\[-2pt]\mathbf{h}_t^k
&= LN (\sum_{e}\pi_{t,e}^k\mathbf{h}_t^e),
\qquad k\in\{\mathit{f},\mathit{w}\},
\end{aligned}
\label{eq:routing}
\end{equation}
where softmax normalizes weights over experts and $LN(\cdot)$ denotes the layer normalization.

For each target $k\in\{\mathit{f},\mathit{w}\}$, we compute attention weights over the valid frames $\mathcal{V}$ and aggregate each routed sequence into clip-level representations:
\begin{equation}
\begin{aligned}
\mathbf{v}^k = \sum_{t\in\mathcal{V}} \beta_t^k\mathbf{h}_t^k; \qquad 
\beta_t^k = \frac{
\exp\!\left(\psi_k(\mathbf{h}_t^k)\right)
}{
\sum_{s\in\mathcal{V}}
\exp\!\left(\psi_k(\mathbf{h}_s^k)\right)
},
\end{aligned}
\label{eq:temporal_pooling}
\end{equation}
where $\psi_k$ predicts a scalar attention score. The representations $\mathbf{v}^{\mathit{f}}$ and $\mathbf{v}^{\mathit{w}}$ are passed to separate prediction heads.

\subsection{Prediction and Learning Objectives}

Two independent regression heads map the target-specific representations $\mathbf{v}^{\mathit{f}}$ and $\mathbf{v}^{\mathit{w}}$ to non-negative estimates of peak contact force $\hat{P}_{\mathit{f}}$ and mechanical work $\hat{P}_{\mathit{w}}$, respectively. During training, the primary loss is defined as $\mathcal{L}_{\mathrm{reg}}=\frac{1}{B}\sum_{i=1}^{B}\left(\frac{|\hat P_{\mathit f,i}-P_{\mathit f,i}|}{s_f}+\frac{|\hat P_{\mathit w,i}-P_{\mathit w,i}|}{s_w}\right)$. The scales $s_f$ and $s_w$ are the 75th percentiles of positive training targets, with lower bounds of $1$~N and $0.05$~J, respectively. We also introduce two frame-level auxiliary losses. The masked Smooth L1 loss $\mathcal{L}_{\mathrm{mean}}$ supervises the frame-level mean force, while a masked binary cross-entropy loss $\mathcal{L}_{\mathrm{env}}$ supervises the loaded interval from the first to the last contact. The overall objective is
\begin{equation}
\mathcal{L}
=\mathcal{L}_{\mathrm{reg}}
+ \lambda_{1}\mathcal{L}_{\mathrm{mean}}
+ \lambda_{2}\mathcal{L}_{\mathrm{env}},
\end{equation}
where both $\lambda_{1}$ and $\lambda_{2}$ are set to 0.06. 

\iffalse
Two independent regression heads map the target-specific representations $\mathbf{v}^{\mathit{f}}$ and $\mathbf{v}^{\mathit{w}}$ to non-negative estimates of peak contact force $\hat{P}_{\mathit{f}}$ and mechanical work $\hat{P}_{\mathit{w}}$, respectively. During training, the primary loss is defined as $\mathcal{L}_{\mathrm{reg}}=\frac{1}{B}\sum_{i=1}^{B}
\left(\frac{|\hat P_{f,i}-P_{f,i}|}{s_f}
+\frac{|\hat P_{w,i}-P_{w,i}|}{s_w}\right)$ consists of the scale-normalized absolute errors of the force and work predictions, with the two targets weighted equally. Target-specific scales are estimated only from the training split to balance their numerical ranges. We also introduce two frame-level auxiliary losses. The masked Smooth L1 loss $\mathcal{L}_{\mathrm{mean}}$ supervises the frame-level mean force, while a masked binary cross-entropy loss $\mathcal{L}_{\mathrm{env}}$ supervises the loaded interval from the first to the last contact. The overall objective is
\begin{equation}
\mathcal{L}
=\mathcal{L}_{\mathrm{reg}}
+ \lambda_{1}\mathcal{L}_{\mathrm{mean}}
+ \lambda_{2}\mathcal{L}_{\mathrm{env}},
\end{equation}
where both $\lambda_{1}$ and $\lambda_{2}$ are set to 0.06. 
\fi

\section{Experiments}

\subsection{Experimental Setup}

\noindent\textbf{Dataset.}
%Following the official Hoi! dataset \cite{engelbracht2026hoi} processing pipeline and manual verification of interaction boundaries, we obtain 336 clips from 22 gripper recordings across 18 scenes. We split these clips into training, validation, and test sets containing 264, 36, and 36 clips, respectively. Each clip contains head-mounted RGB video synchronized with gripper force/torque and end-effector motion measurements, from which we derive labels for peak contact force and contact-phase mechanical work. 
Following the Hoi! processing pipeline \cite{engelbracht2026hoi} and manual verification of interaction boundaries, we obtain 336 clips from 22 gripper recordings across 18 scenes. We randomly split the 336 clips into 264 training, 36 validation, and 36 test clips for experiments. Each clip pairs head-mounted RGB video with synchronized gripper force/torque and end-effector motion measurements, from which we derive peak contact force and contact-phase mechanical work labels.

%We randomly select one opening and one closing clip per scene for a fixed 36-clip test set, and split the remaining clips into 264 for training and 36 for validation

\noindent\textbf{Evaluation metrics.}
We evaluate predictions using mean absolute error (MAE), median absolute error (MedianAE), root mean squared error (RMSE), and mean relative error (MRE). MAE, MedianAE, and RMSE are reported in newtons for force and joules for work. MRE is defined as:

\begin{equation}
\mathrm{MRE}
=\frac{1}{M}\sum_{i=1}^{M}
\frac{|\hat{P}_i-P_i|}{\max(|P_i|,10^{-8})},
\end{equation}
%\vspace{-0.2cm}
where $M$ is the number of test clips. 

\begin{table}[t]
\centering
\caption{Comparison with baseline methods on test split.}
\small
\label{tab:main_results}
\scriptsize
\setlength{\tabcolsep}{2.0pt}
\renewcommand{\arraystretch}{1.08}
\resizebox{\columnwidth}{!}{%
\begin{tabular}{lrrrr}
\toprule
\multicolumn{5}{c}{\textbf{Peak Contact Force} } \\
\cmidrule(lr){1-5}
Methods & MAE ($N$) $\downarrow$ & MedianAE ($N$) $\downarrow$ & RMSE ($N$) $\downarrow$ & MRE $\downarrow$ \\
\midrule
DINOv3-only
& 9.320 $\pm$ 0.189
& 5.696 $\pm$ 0.337
& 13.827 $\pm$ 0.290
& 0.584 $\pm$ 0.015 \\
DA3-only
& 9.318 $\pm$ 0.770
& 5.702 $\pm$ 0.523
& 13.689 $\pm$ 1.334
& 0.596 $\pm$ 0.039 \\
DINOv3+DA3
& 8.796 $\pm$ 1.230
& 5.702 $\pm$ 0.801
& 12.689 $\pm$ 1.949
& 0.600 $\pm$ 0.068 \\
\rowcolor{gray!15}
Full Model
& \textbf{5.205 $\pm$ 0.584}
& \textbf{3.492 $\pm$ 0.594}
& \textbf{7.132 $\pm$ 0.765}
& \textbf{0.301 $\pm$ 0.031} \\
%\addlinespace[2pt]
\midrule
\multicolumn{5}{c}{\textbf{Mechanical Work} } \\
\cmidrule(lr){1-5}
Methods & MAE ($J$) $\downarrow$  & MedianAE ($J$) $\downarrow$ & RMSE ($J$) $\downarrow$ & MRE $\downarrow$ \\
\midrule
DINOv3-only
& 1.700 $\pm$ 0.032
& 0.665 $\pm$ 0.139
& 3.088 $\pm$ 0.094
& 1.948 $\pm$ 0.347 \\
DA3-only
& 1.656 $\pm$ 0.128
& 0.823 $\pm$ 0.037
& 2.751 $\pm$ 0.234
& 2.146 $\pm$ 0.265 \\
DINOv3+DA3
& 1.624 $\pm$ 0.124
& 0.852 $\pm$ 0.050
& 2.695 $\pm$ 0.226
& 2.034 $\pm$ 0.337 \\
\rowcolor{gray!15}
Full Model
& \textbf{0.894 $\pm$ 0.081}
& \textbf{0.427 $\pm$ 0.045}
& \textbf{1.508 $\pm$ 0.203}
& \textbf{1.237 $\pm$ 0.189} \\
\bottomrule
\end{tabular}%
}
%\vspace{-0.2cm}
\end{table}

% ===== E4 single-sample case figure (single column) =====
%\vspace{-0.3cm}
\begin{figure}[t]
    \centering
    \includegraphics[width=\linewidth]{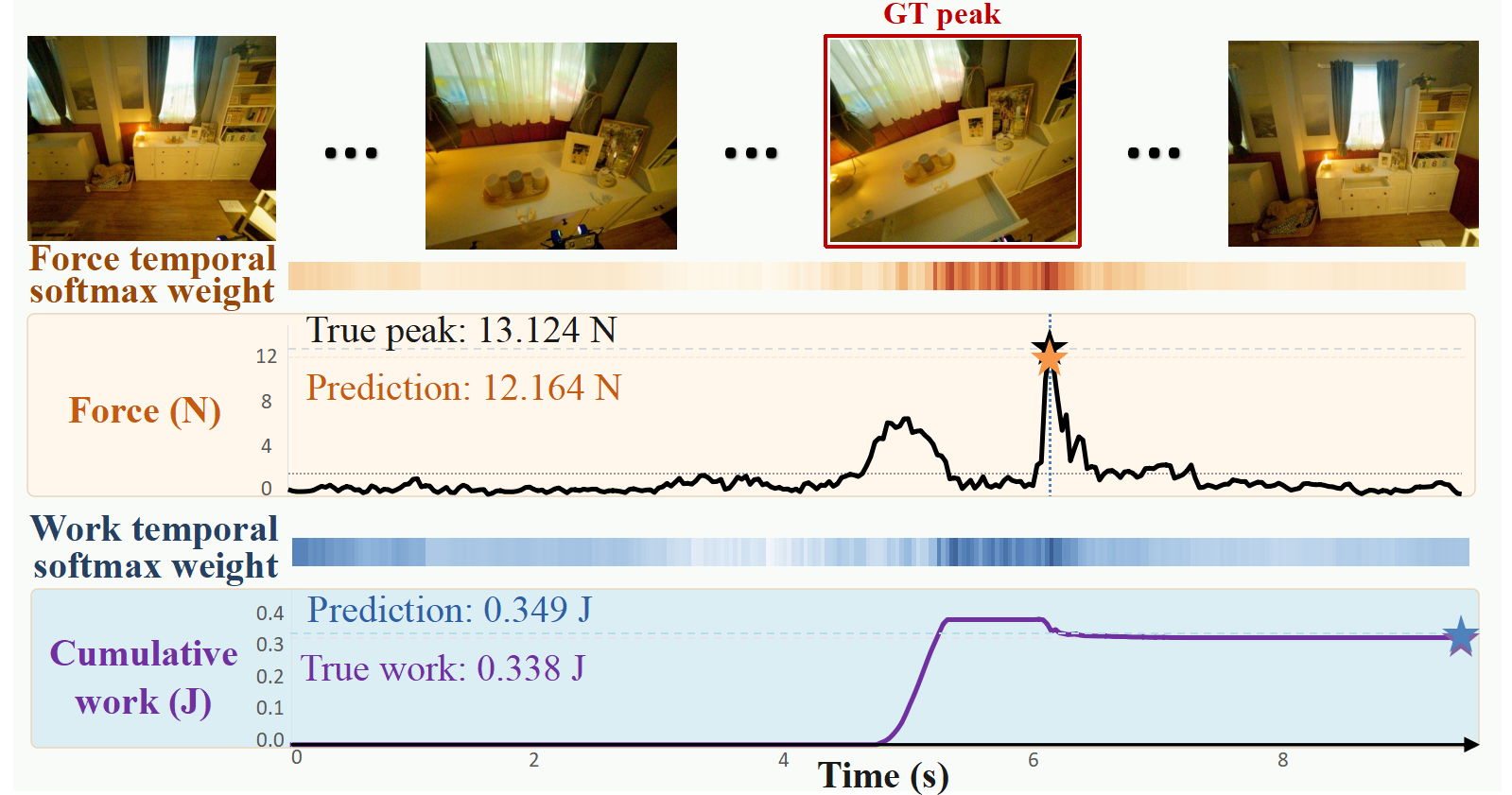}
    \vspace{-0.6cm}
    \caption{Qualitative results on the selected test video clip. }
    \label{fig:e4_case_summary}
    \vspace{-0.6cm}
\end{figure}

\noindent\textbf{Implementation details.}
We extract appearance and geometry features offline using frozen DINOv3 and DA3 encoders, respectively, and pool their dense feature maps into $7\times7$ grids. The common embedding dimension $D$ and the hidden dimension of each temporal expert are both set to $192$. We train EgoPhys for 60 epochs using AdamW with a batch size of 2 and an initial learning rate of $2\times10^{-4}$. Results are reported as the mean and standard deviation over six random seeds.

%We extract appearance and geometry features offline using frozen DINOv3 and DA3 encoders, respectively, and pool their feature maps to a $7\times7$ spatial grid. The hidden dimension of each temporal expert is set to $192$. We train EgoPhys for 60 epochs using AdamW with a batch size of 2 and an initial learning rate of $2\times10^{-4}$. Results are reported as the mean and sample standard deviation over six random seeds.

\subsection{Experimental Results}

We compare EgoPhys with three baselines: DINOv3-only and DA3-only apply spatial average pooling to the appearance and geometry features from their respective encoders, while DINOv3+DA3 concatenates the two pooled representations at each frame. All baselines use temporal attention pooling to obtain clip-level representations, followed by separate prediction heads for the two targets. As shown in Table \ref{tab:main_results}, EgoPhys achieves the lowest errors across all four metrics for both force and work. Compared with DINOv3+DA3, it reduces the force and work MAEs from 8.796 $N$ to 5.205 $N$ and from 1.624 $J$ to 0.894 $J$, respectively. Fig. \ref{fig:e4_case_summary} presents one test example in which both predictions are close to their ground truth values. The force branch concentrates its temporal weights near the force peak, whereas the work branch distributes them more broadly over the interaction, illustrating their different temporal dependencies.

%We compare EgoPhys with three baselines: DINOv3-only and DA3-only apply spatial average pooling to the appearance and geometry features from their respective encoders, while DINOv3+DA3 concatenates the two pooled representations at each frame. All baselines use temporal attention pooling to obtain clip-level representations, followed by separate prediction heads for predictions. As shown in Table \ref{tab:main_results}, EgoPhys achieves the lowest errors across all four metrics for both force and work. The MAE of force and work are $5.205$~N and $0.894$~J, respectively. Fig.~\ref{fig:e4_case_summary} illustrates a held-out clip with predictions close to ground truth for both targets.

%We compare EgoPhys with three baselines: DINO-only, DA3-only, and DINOv3+DA3. DINO-only and DA3-only apply spatial average pooling to the appearance and geometry features from their respective encoders, while DINOv3+DA3 concatenates the two pooled representations at each frame. All baselines use temporal attention pooling to obtain clip-level representations, followed by separate prediction heads for peak contact force and mechanical work. As shown in Table~\ref{tab:main_results}, EgoPhys achieves the lowest errors across all four metrics for both targets. Its force and work MAEs are $5.205$~N and $0.894$~J, respectively, representing reductions of $40.8\%$ and $45.0\%$ over DINOv3+DA3, the strongest baseline in MAE. Fig.~\ref{fig:e4_case_summary} illustrates a held-out clip with predictions close to the ground truth for both targets.

\subsection{Ablation Study}

\begin{table}[t]
\centering
\caption{Ablation of the main components.}
\label{tab:ablation-designs}
\scriptsize
\setlength{\tabcolsep}{2.0pt}
\renewcommand{\arraystretch}{1.08}
\resizebox{\columnwidth}{!}{%
\begin{tabular}{@{}lrrrr@{}}
\toprule
& \multicolumn{2}{c}{\textbf{Peak Contact Force}}
& \multicolumn{2}{c@{}}{\textbf{Mechanical Work}} \\
\cmidrule(lr){2-3}\cmidrule(l){4-5}
Models
& MAE ($N$) $\downarrow$
& MRE $\downarrow$
& MAE ($J$) $\downarrow$
& MRE $\downarrow$ \\
\midrule
\#A
& 8.796 $\pm$ 1.230
& 0.600 $\pm$ 0.068
& 1.624 $\pm$ 0.124
& 2.034 $\pm$ 0.337 \\

\#B
& 7.776 $\pm$ 0.308
& 0.447 $\pm$ 0.024
& 1.456 $\pm$ 0.210
& 2.062 $\pm$ 0.476 \\

\#C
& 5.982 $\pm$ 0.299
& 0.335 $\pm$ 0.030
& 1.103 $\pm$ 0.169
& 1.428 $\pm$ 0.222 \\

\rowcolor{gray!15}
Full model
& \textbf{5.205 $\pm$ 0.584}
& \textbf{0.301 $\pm$ 0.031}
& \textbf{0.894 $\pm$ 0.081}
& \textbf{1.237 $\pm$ 0.189} \\
\bottomrule
\end{tabular}%
}
\vspace{-0.4cm}
\end{table}

\begin{table}[t]
\centering
\caption{Comparison of temporal expert configurations.}
\label{tab:expert-comparison}
\scriptsize
\setlength{\tabcolsep}{2.0pt}
\renewcommand{\arraystretch}{1.08}
\resizebox{\columnwidth}{!}{%
\begin{tabular}{@{}lrrrr@{}}
\toprule
& \multicolumn{2}{c}{\textbf{Peak Contact Force}}
& \multicolumn{2}{c@{}}{\textbf{Mechanical Work}} \\
\cmidrule(lr){2-3}\cmidrule(l){4-5}
Expert
& MAE ($N$) $\downarrow$
& MRE $\downarrow$
& MAE ($J$) $\downarrow$
& MRE $\downarrow$ \\
\midrule
Semantic
& 5.753 $\pm$ 0.267
& 0.319 $\pm$ 0.018
& 1.093 $\pm$ 0.074
& 1.467 $\pm$ 0.219 \\

Event
& 6.401 $\pm$ 0.396
& 0.346 $\pm$ 0.036
& 1.344 $\pm$ 0.172
& 1.801 $\pm$ 0.269 \\

Motion
& 7.039 $\pm$ 0.570
& 0.432 $\pm$ 0.043
& 1.571 $\pm$ 0.067
& 1.883 $\pm$ 0.314 \\

All (Concat)
& 6.303 $\pm$ 0.263
& 0.359 $\pm$ 0.023
& 1.189 $\pm$ 0.037
& 1.578 $\pm$ 0.232 \\

\rowcolor{gray!15}
All (Gate)
& \textbf{5.205 $\pm$ 0.584}
& \textbf{0.301 $\pm$ 0.031}
& \textbf{0.894 $\pm$ 0.081}
& \textbf{1.237 $\pm$ 0.189} \\
\bottomrule
\end{tabular}%
}
\vspace{-0.4cm}
\end{table}

Table~\ref{tab:ablation-designs} evaluates the main components for peak contact force and mechanical work prediction. Adding CASA (\#B) to the DINOv3+DA3 baseline (\#A) reduces the respective MAEs from 8.796 $N$ to 7.776 $N$ and from 1.624 $J$ to 1.456 $J$, suggesting that spatially localized interaction cues benefit both targets. TMTR (\#C) further reduces the MAEs to 5.982 $N$ and 1.103 $J$, indicating the value of modeling temporal evidence beyond frame-level aggregation. The auxiliary mean-force and loaded-envelope losses further reduce the force and work MAEs to 5.205 $N$ and 0.894 $J$, achieving the lowest errors for peak contact force and mechanical work. As shown in Table~\ref{tab:expert-comparison}, the semantic expert is the strongest individual expert, yet direct concatenation of all three experts performs worse. The full gated configuration improves upon concatenation for both force (5.205 $N$ versus 6.303 $N$) and work (0.894 $J$ versus 1.189 $J$). These experiments suggest that semantic context is informative on its own, while combining temporal cues selectively is more effective than simply concatenating them.

\section{Conclusion}

We presented EgoPhys, a framework for jointly estimating peak contact force and contact-phase mechanical work from egocentric RGB video. EgoPhys combines CASA for aggregating interaction-relevant appearance and geometry evidence with TMTR for modeling semantic, event, and motion cues through target-specific routing. On the fixed test split constructed from the Hoi! dataset, EgoPhys achieves MAEs of $5.205\pm0.584$ $N$ for peak contact force and $0.894 \pm 0.081$ $J$ for mechanical work, representing significant improvements over the DINOv3+DA3 baseline. Ablation studies further demonstrate the effectiveness of CASA and TMTR modules. In future work, we will further investigate the generalization of EgoPhys to unseen scenes and objects in egocentric videos.

%We presented an RGB-only framework for jointly predicting peak contact force and contact mechanical work from egocentric clips. The model combines interaction-centered appearance--geometry fusion, complementary semantic, event, and motion temporal experts, and independent target-specific routing, with physical signals confined to training-time supervision. The progressive ablation results favor the complete design over simpler variants and are consistent with the need for distinct temporal evidence for the two targets. Because the evaluation uses a small sample-level outer test, future work should emphasize larger scene-held-out benchmarks and broader physical supervision.

\bibliographystyle{IEEEbib}
%\small
\bibliography{refs}

\end{document}